%% file: main.tex
\documentclass[letterpaper, 10 pt, conference]{ieeeconf}  
\include{preambles}

\include{commands}
\IEEEoverridecommandlockouts 
\title{\LARGE \bf
Egocentric Vision-based Future Vehicle Localization \\
for Intelligent Driving Assistance Systems
}

\author{Yu Yao$^{1*}$, Mingze Xu$^{2*}$, Chiho Choi$^{3}$, David J. Crandall$^{2}$, Ella M. Atkins$^{1}$, Behzad Dariush$^{3}$% <-this % stops a space
\thanks{$^{1}$Robotics Institute, University of Michigan, Ann Arbor, MI 48109, USA.
{\tt\footnotesize \{brianyao,ematkins\}@umich.edu}}%
\thanks{$^{2}$School of Informatics, Computing and Engineering, Indiana University, Bloomington, IN 47408, USA. 
{\tt\footnotesize \{mx6,djcran\}@indiana.edu}}%
\thanks{$^{3}$Honda Research Institute, Mountain View, CA 94043, USA.
{\tt\footnotesize \{cchoi, bdariush\}@honda-ri.com}}%
\thanks{$^{*}$This work was done when Yu Yao and Mingze Xu were interns at Honda Research Institute, Mountain View, CA 94043, USA.}%
}
\begin{document}
\maketitle
\thispagestyle{empty}
\pagestyle{empty}

\begin{abstract}

Predicting the future location of vehicles is essential for
safety-critical applications such as advanced driver assistance
systems (ADAS) and autonomous driving.  This paper introduces a novel
approach to simultaneously predict both the location and scale
of target vehicles in the first-person (egocentric) view
of an ego-vehicle.
We present a multi-stream recurrent neural network (RNN)
encoder-decoder model that separately captures both object location and scale
and pixel-level observations for future vehicle localization. We show that incorporating dense optical flow improves
prediction results significantly since it captures information about
motion as well as appearance change. We also find that 
explicitly modeling future motion of the ego-vehicle improves the prediction accuracy,
which could be especially beneficial in
intelligent and automated vehicles that have motion planning
capability. To evaluate the performance of our approach, we present
a new dataset of first-person videos collected from a variety of
scenarios at road intersections, which are particularly challenging moments
for prediction because vehicle trajectories are diverse and dynamic. \textit{Code and dataset have been made available at: \url{https://usa.honda-ri.com/hevi}}

\end{abstract}

\input{Introduction}
\input{RelatedWork}
\input{Method}
\input{Experiments}
\input{Conclusion}

\clearpage
\bibliographystyle{IEEEtran}
\bibliography{Reference}

\end{document}

%% file: preambles.tex
\usepackage{amsmath} % assumes amsmath package installed
\usepackage{amssymb}  % assumes amsmath package installed
\usepackage{amsfonts} % assumes amsfonts package installed
\usepackage[utf8]{inputenc}

\usepackage[pdftex]{graphicx}
\usepackage[usenames, dvipsnames]{color}
\usepackage{color,soul}
\setstcolor{red}

\usepackage[labelformat=simple]{subcaption}

\usepackage{multirow}
\usepackage{makecell}
\usepackage[linesnumbered,ruled]{algorithm2e}

\usepackage{array}
\usepackage{url}

\usepackage{nomencl}

\usepackage{amsthm}
\usepackage{tikz}

\theoremstyle{definition}

\theoremstyle{remark}

\usepackage{makecell}
\usepackage{amssymb}

\usepackage{lipsum}% http://ctan.org/pkg/lipsum
\usepackage{multicol}% http://ctan.org/pkg/multicols

\usepackage{bm}

\usepackage{hyperref}
\hypersetup{
    colorlinks=true,
    linkcolor=blue,
    filecolor=magenta,      
    urlcolor=magenta,
}
\usepackage{relsize}
\usepackage{cleveref}
\usepackage{epsfig}
\usepackage{graphicx}
\usepackage{array}
\usepackage{booktabs}
\usepackage{xcolor}
\usepackage{cite}
\usepackage{multirow}

%% file: commands.tex
\newcommand{\xhdr}[1]{\vspace{6pt} \noindent {\textbf{#1} }}
\newcommand{\etal}{\textit{et al}.}

\newcommand{\RNum}[1]{\uppercase\expandafter{\romannumeral #1\relax}}

\usepackage{etoolbox}
\makeatletter
\patchcmd{\@makecaption}
  {\scshape}
  {}
  {}
  {}
\makeatother

%% file: Introduction.tex
\section{Introduction}

Safe driving requires not just accurately identifying and locating nearby
objects, but also predicting their \textit{future} locations and actions so
that there is enough time to avoid collisions. Precise
prediction of nearby vehicles' future locations is thus essential for
both autonomous and semi-autonomous (e.g., Advanced Driver
Assistance Systems, or ADAS) driving systems as well as safety-related systems~\cite{yao2018smart}. 
Extensive research~\cite{Deo2018,deo2018convolutional,lee2017desire}
has been conducted on predicting vehicles' future actions and
trajectories using overhead (bird's eye view) observations.  
But obtaining overhead views requires either an externally-mounted camera (or LiDAR),
which is not common on today's production vehicles,
or aerial imagery that must be transfered to the vehicle over a network connection.

\begin{figure}[tbp]
    \centering
    \includegraphics[width=1\columnwidth]{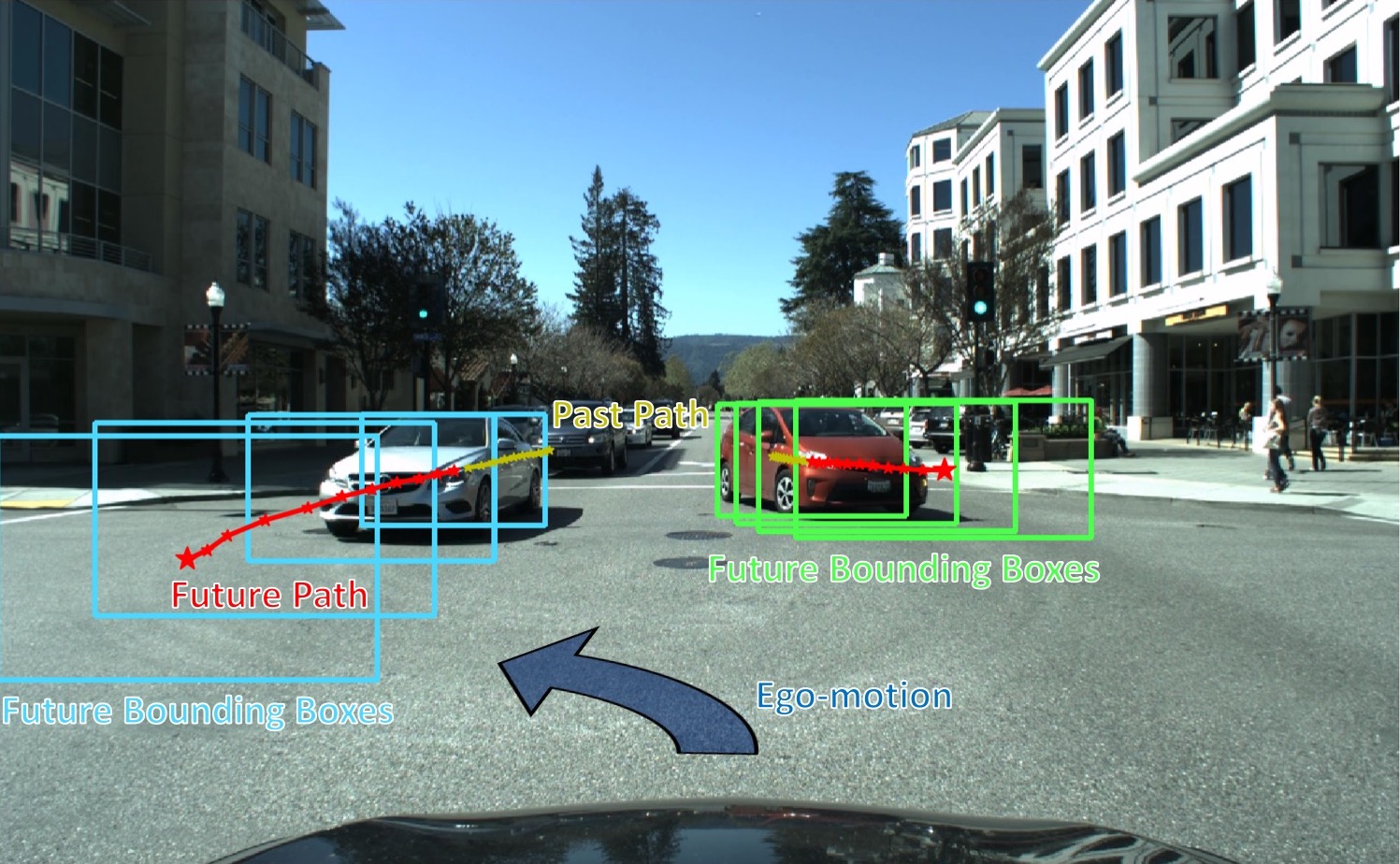}
    \caption{Illustration of future vehicle localization. Location and scale
    are represented as bounding boxes in predictions.}
    \vspace{-10pt}
    \label{fig:intro}
\end{figure}

A much more natural approach is to use forward-facing cameras that
record the driver's ``first-person'' or ``egocentric'' perspective.
In addition to being easier to collect, the first-person perspective captures
rich information about the object appearance, as well as the relationships and interactions between the
ego-vehicle and objects in the environment. Due to these advantages,
egocentric videos have been directly used in applications such
as action recognition~\cite{li2015delving,ma2016going},
navigation~\cite{anderson2017vision,das2017embodied,ye2018active}, and end-to-end
autonomous driving~\cite{codevilla2017end}.
For trajectory prediction, some work has simulated bird's eye views by
projecting egocentric video frames onto the ground plane~\cite{Deo2018,deo2018convolutional}, but these
projections can be incorrect due to road irregularities or other sources
of distortion, which prevent accurate vehicle position prediction. 

This paper considers the challenging problem of predicting relative future
locations and scales (represented as bounding boxes in Figure~\ref{fig:intro}) of nearby
vehicles with respect to an ego-vehicle equipped with an egocentric
camera. We introduce a multi-stream RNN encoder-decoder (RNN-ED) architecture
to effectively encode past observations from different domains and
generate future bounding boxes. Unlike
other work that has addressed prediction in simple scenarios such as
freeways~\cite{Deo2018,deo2018convolutional}, we consider
urban driving scenarios with a variety of multi-vehicle
behaviors and interactions.

The contributions of this paper are three-fold.
First, our work presents a novel perspective for intelligent driving systems
to predict vehicle's future location under egocentric
view and challenging driving scenarios such as intersections.
Second, we propose a multi-stream RNN-ED architecture for improved
temporal modeling and explicitly capturing vehicles' motion as well as
their appearance information by using dense optical flow and
future ego-motion as inputs.
Third, we publish a new first-person video
dataset --- the Honda Egocentric View - Intersection (HEV-I) dataset
--- collected in a variety of scenarios involving road intersections. The dataset includes over $2,400$
vehicles (after filtering) in $230$ videos.
We evaluate our approach on this new proposed dataset,
along with the existing KITTI dataset, and achieve the state-of-the-art results.

%% file: RelatedWork.tex
\section{Related Work}\label{sec:related_work}

\xhdr{Egocentric Vision.}  An egocentric camera view is
often the most natural perspective for observing an ego-vehicle
environment, but it introduces additional challenges due to its narrow
field of view.  The literature in egocentric visual perception has
typically focused on activity
recognition~\cite{fathi2011understanding,kitani2011fast,pirsiavash2012detecting,li2015delving,ma2016going},
object detection~\cite{lee2015predicting,bertasius2016first,gao2019egocentric}, person
identification~\cite{ardeshir2016ego2top,fan2017identifying,xu2018joint},
video summarization~\cite{lee2012discovering}, and gaze
anticipation~\cite{li2013learning}.  Recently, papers have also
applied egocentric vision to ego-action estimation and prediction.
For example, Park \etal~\cite{soo2016egocentric} proposed a method to
estimate the location of a camera wearer in future video frames.  Su
\etal~\cite{su2017predicting} introduced a Siamese network to
predict future behaviors of basketball players in multiple
synchronized first-person views. Bertasius \etal~\cite{bertasius2018egocentric}
addressed the motion planning problem for generating an egocentric
basketball motion sequence in the form of a 12-d camera configuration
trajectory.

More directly related to our problem, two recent papers have
considered predicting pedestrians' future locations from egocentric
views. Bhattacharyya \etal~\cite{Bhattacharyya_2018_CVPR} model
observation uncertainty using Bayesian Long Short-Term Memory (LSTM)
networks to predict the distribution of possible future locations.
Their technique does not try incorporate image features such as object appearance.
Yagi \etal~\cite{Yagi_2018_CVPR} use human pose, scale,
and ego-motion as cues in a convolution-deconvolution (Conv1D)
framework to predict future locations. The specific pose information applies to people but not to other on-road objects like vehicles. Their
Conv1D model captures important features of the activity sequences but
does not explicitly model temporal updating along each
trajectory. In contrast, our paper proposes a multi-stream RNN-ED
architecture using past vehicle locations and image features as
inputs for predicting vehicle locations from egocentric views.

\xhdr{Trajectory Prediction.}  Previous work on vehicle
trajectory prediction has used motion features and probabilistic
models~\cite{Deo2018, wiest2012probabilistic}. The probability of
specific motions (e.g., lane change) is first estimated, and the
future trajectory is predicted using Kalman filtering. Computer vision
and deep learning techniques achieved convincing results in
several fields~\cite{He2017,Ilg2017,xu2018temporal}, and have been recently investigated
for trajectory prediction. Alahi
\etal~\cite{Alahi_2016_CVPR} proposed Social-LSTM to model pedestrian
trajectories as well as their interactions. The proposed social pooling
method was then improved by Gupta \etal~\cite{gupta2018social} to
capture global context for a Generative Adversarial Network
(GAN). Social pooling is first applied to vehicle trajectory
prediction in Deo \etal~\cite{deo2018convolutional} with multi-modal
maneuver conditions. Other work models scene context information 
using attention mechanisms to assist trajectory
prediction~\cite{sadeghian2018car,sadeghian2018sophie}.  Lee
\etal~\cite{lee2017desire} incorporate RNN models with conditional
variational autoencoders to generate multimodal predictions, and
select the best prediction by ranking scores. 

However, these methods model trajectories and context information
from a bird's eye view in a static camera setting, which significantly
simplifies the challenge of measuring distance from visual features.  In contrast, in monocular
first-person views, physical distance can be estimated only indirectly, through scaling
and observations of participant vehicles, and the environment changes
dynamically due to ego-motion effects.  Consequently, previous work
cannot be directly applied to first-person videos.  On the other hand, the first-person view provides
higher quality object appearance information compared to birds eye view images, in 
which objects are represented only by the coordinates of their
geometric centers. This paper encodes past location, scale, and
corresponding optical flow fields of target vehicles to predict their
future locations, and  we further improve prediction performance by incorporating future ego-motion.

%% file: Method.tex
\section{Future vehicle localization from \\
first-person views}\label{sec:method}

We now present our approach to predicting future bounding boxes of vehicles in first-person view.
Our method differs from traditional trajectory
prediction because the distances of object motion in perspective
images do not correspond to physical distances directly, and because the motion of the camera
(ego-motion) induces additional apparent motion on nearby objects.

Consider a vehicle visible in the egocentric field of view, 
and let its past
bounding box trajectory be
$\mathbf{X}=\{X_{t_0-\tau+1},X_{t_0-\tau+2},...,X_{t_0}\}$, where
$X_t=[c^x_t,c^y_t,w_t,h_t]$ is the bounding box of the vehicle at 
time $t$ (i.e., its center location and width and height in pixels, respectively).
Similarly, let the future bounding box trajectory be given by
$\mathbf{Y}=\{Y_{t_0+1},Y_{t_0+2},...,Y_{t_0+\delta}\}$.  Given
image evidence observed from the past $\tau$ frames,
$\mathbf{O}=\{O_{t_0-\tau+1},O_{t_0-\tau+2},...,O_{t_0}\}$, and its corresponding
past bounding box trajectory $\mathbf{X}$,
our goal is to predict
$\mathbf{Y}$. 

\begin{figure*}
    \vspace{5pt}
    \centering
    \includegraphics[width = 0.92\textwidth]{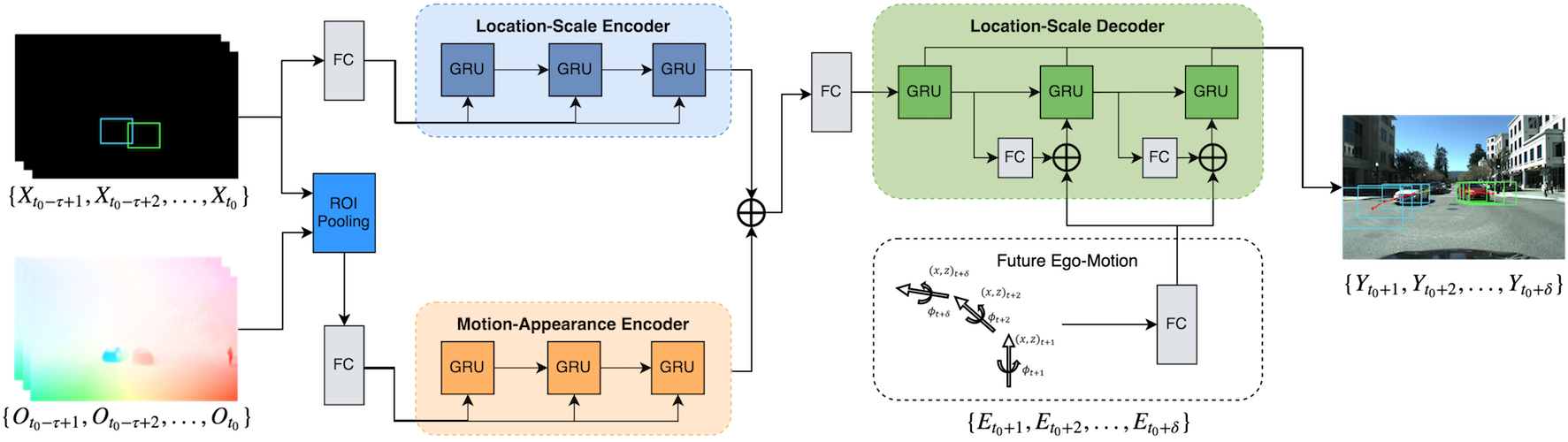}
    \caption{The proposed 
    future vehicle localization framework (better in color).}
    \label{fig:RNN-ED}
\end{figure*}

We propose a multi-stream RNN encoder-decoder (RNN-ED) model
to encode temporal information of past observations
and decode future bounding boxes, as shown in Figure~\ref{fig:RNN-ED}.
The past bounding box trajectory is
encoded to provide location and scale information, while dense
optical flow is encoded to provide pixel-level information about vehicle
scale, motion, and appearance changes.
Our decoder can also consider
information about future ego-motion, which could be available
from the planner of an intelligent vehicle.
The decoder generates hypothesized future bounding boxes by temporally
updating from the encoded hidden state.

\subsection{Temporal Modeling}
\subsubsection{Location-Scale Encoding}
One straightforward approach
to predict the future location of an object is to extrapolate a future trajectory from the past. However, in
perspective images, physical object location is reflected by both its
pixel location and scale. For example, a vehicle located at the
center of an image could be a nearby lead vehicle or a distant vehicle
across the intersection, and such a difference could cause a completely
different future motion.  Therefore, this paper predicts both the
location and scale of participant vehicles, i.e., their bounding boxes. The scale information
is also able to represent depth (distance) as well as vehicle orientation, given that
distant vehicles tend to have smaller bounding boxes and crossing
vehicles tend to have larger aspect ratios.

\subsubsection{Motion-Appearance Encoding}

Another important cue for predicting a vehicle's future location is pixel-level information about motion and
appearance. Optical flow is widely used as a pattern of relative
motion in a scene. For each feature point, optical flow gives an estimate of a vector $[u,v]$ that
describes its relative motion from one frame to the next caused by the motion of the
object and the camera. Compared to sparse optical flow obtained from
traditional methods such as Lucas-Kanade \cite{lucas1981iterative}, dense optical flow offers
an estimate at every pixel, so that moving objects can be
distinguished from the background. Also, dense optical flow captures
object appearance changes, since different object pixels may have
different flows,
as shown in the left part of
Fig.~\ref{fig:RNN-ED}.

In this paper, object vehicle features are extracted by a region-of-interest pooling (ROIPooling)
operation using bilinear interpolation from the optical flow map. The
ROI region is expanded from the bounding box to contain contextual
information around the object, so that its relative motion with respect
to the environment is also encoded. The resulting relative motion
vector is represented as $O_t = [u_1, v_1, u_2, v_2,...u_n,v_n]_t$,
where $n$ is the size of the pooled region.  

We use two encoders for temporal modeling of
each input stream and apply the late fusion method:
\begin{subequations}
    \begin{align}
        h^X_{t} & = GRU_{_X}(\phi_{_X}(X_{t-1}), h^X_{t-1}; \theta_{_X})\\
        h^O_{t} &  = GRU_{_O}(\phi_{_O}(O_{t-1}),h^O_{t-1}; \theta_{_O})\\
        \mathcal{H} & = \phi_{\mathcal{H}}(Average(h^X_{t_0}, h^O_{t_0})) 
    \end{align}
\end{subequations}
\noindent where $GRU$ represents the gated recurrent units~\cite{cho2014learning} with parameter $\theta$, $\phi(\cdot)$ are
linear projections with ReLU activations, and $h^x_{t}$ and $h^o_{t}$
are the hidden state vectors of the GRU models at time $t$.

\subsection{Future Ego-Motion Cue}
Awareness of future ego-motion is essential to predicting the future location
of participant vehicles. For autonomous vehicles, it is reasonable to
assume that motion planning (e.g. trajectory generation) is available \cite{gonzalez2016review},
so that the future pose of the ego
vehicle can be used to aid in predicting the relative position
of nearby vehicles. Planned ego-vehicle motion information may
also help anticipate motion caused by interactions between vehicles:
the ego-vehicle turning left at intersection may result in other vehicles stopping to yield or accelerating to pass, for example.

In this paper, %\st{ground is flat, and thus only must consider 2D transformations}. 
the future ego motion is represented by 2D rotation matrices
$R_{t}^{t+1}\in\mathbb{R}^{2\times 2}$ and translation vectors
$T_{t}^{t+1}\in\mathbb{R}^2$ \cite{Yagi_2018_CVPR}, which together describe the transformation of the camera coordinate frame from time $t$ to
$t+1$. The relative, pairwise transformations between frames can be composed to estimate transformations across the prediction horizon from the current frame:
\begin{subequations}
    \begin{equation}
        % R_0^{t_0+i} = \prod_{j=1}^{t_0+i}R_{j-1}^{j} \\
        R_{t_0}^{t_0+i} = \prod_{t=t_0}^{t_0+i-1}R_{t}^{t+1}\
    \end{equation}
    \begin{equation}
        T_{t_0}^{t_0+i} = T_{t_0}^{t_0+i-1} + R_{t_0}^{t_0+i-1}T_{t_0+i-1}^{t_0+i}
    \end{equation}
\end{subequations}
The future ego-motion feature is represented by a vector
$E_{t}=[\psi^t_{t_0}, x^t_{t_0},z^t_{t_0}]$, where $t>t_0$,
$\psi^t_{t_0}$ is the yaw angle extracted from $R_{t_0}^t$, and
$x^t_{t_0}$ and $z^t_{t_0}$ are translations from the coordinate frame at time
$t_0$. We use a right-handed coordinate fixed to ego vehicle, where vehicle heading aligns with positive $x$. Estimated future motion is then used as input to the
trajectory decoding model.

\subsection{Future Location-Scale Decoding}

We use another GRU for decoding future bounding boxes. The
decoder hidden state is initialized from the final fused hidden state of the
past bounding box encoder and the optical flow encoder:
\begin{subequations}
	\begin{equation}
	    h^Y_{t+1}  = GRU_{_Y}(f(h^Y_t, E_t), h^Y_t; \theta_{_Y})
	\end{equation}
	\begin{equation}\label{eq:output}
	    Y_{t_0+i}-X_{t_0}=\phi_{out}(h^Y_{t_0+i})
	\end{equation}
	 \begin{equation}
	     f(h^Y_t, E_t)  = Average(\phi_{_Y}(h^Y_t), \phi_e(E_t))
	 \end{equation}
\end{subequations}
\noindent where $h^Y_t$ is the decoder's hidden state, $h^Y_{t_0} =
\mathcal{H}$ is the initial hidden state of the decoder, and $\phi(\cdot)$
are linear projections with ReLU activations applied for domain
transfer. Instead of directly generating the future bounding boxes $\textbf{Y}$, our RNN decoder generates the relative
location and scale of the future bounding box from the current frame
as in \eqref{eq:output}, similar to \cite{Yagi_2018_CVPR}. In this way, the model output is shifted to have zero initial, which improves the performance.

%% file: Experiments.tex
\section{Experiments}\label{sec:experiment}

\subsection{Dataset}

The problem of future vehicle localization in egocentric cameras is
particularly challenging when multiple vehicles execute different motions (e.g.
ego-vehicle is turning left but yields to another moving car).
However, to the best of our knowledge, most existing autonomous
driving datasets are proposed for scene understanding
tasks~\cite{Geiger2013IJRR,Cordts2016Cityscapes} that do not contain much
diverse motion. This paper introduces a new egocentric vision
dataset, the \textit{Honda Egocentric View-Intersection} (HEV-I) data,
that focuses on intersection scenarios where
vehicles exhibit diverse motions due to complex road layouts and
vehicle interactions. HEV-I was collected from different intersection
types in the San Francisco Bay Area, and
consists of $230$ videos each ranging between $10$ to $60$
seconds. Videos were captured by an RGB camera mounted on the
windshield of the car, with $1920 \times 1200$ resolution (reduced to
$1280 \times 640$ in this paper) at $10$ frames per second (fps).

\begin{figure}[t]
    \vspace{5pt}
    \centering
    \includegraphics[width=1\columnwidth]{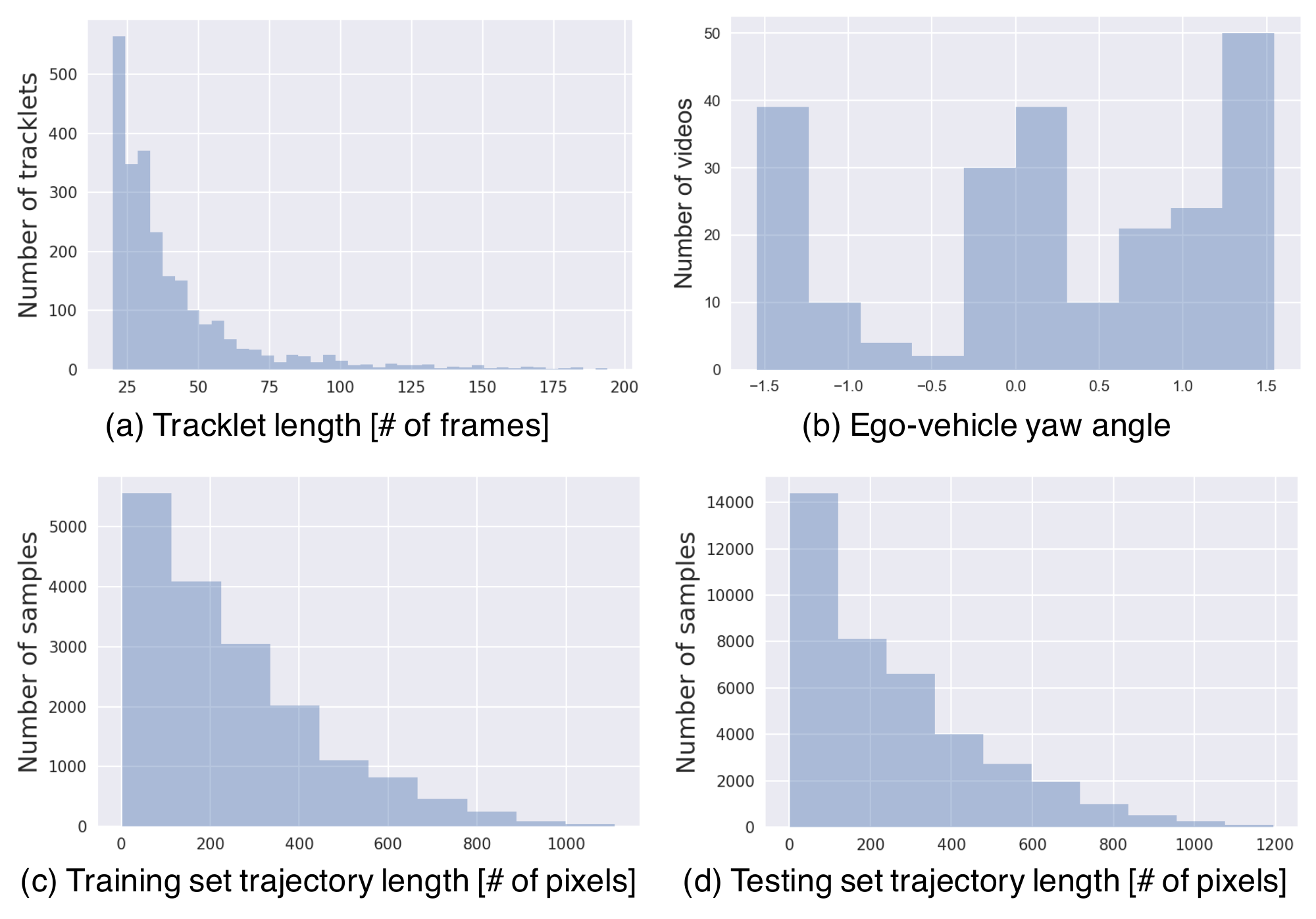}
    \caption{HEV-I dataset statistics.}
    \label{fig:dataset_dist}
    \vspace{-15pt}
\end{figure}

\begin{table}[h]
    \centering
    \caption{Comparison with KITTI dataset. The number of vehicles is tallied after filtering out short sequences.}
    \label{tab:datasets}
    \footnotesize{
    \begin{tabular}{c|c|c|c}
        \toprule
        Dataset & \# videos & \# vehicles & scene types  \\
        \midrule
        KITTI & $38$ & $541$ & residential, highway, city road \\
        % Cityscapes & $~50$ &  & multiple \\
        HEV-I & $230$ & $2477$ & urban intersections\\
        \bottomrule
    \end{tabular}}
\end{table}

Following prior work~\cite{Yagi_2018_CVPR}, we first detected vehicles by using
Mask-RCNN~\cite{He2017} pre-trained on the COCO dataset. We then used Sort~\cite{Bewley2016_sort} with a Kalman
filter for multiple object tracking over each video. In first-person videos, the
duration of vehicles can be extremely short due to high relative
motion and limited fields of view.  On the other
hand, vehicles at stop signs or traffic lights do not move at all
over a short period. In our dataset, we found a sample of $2$ seconds
length is reasonable for including many vehicles while
maintaining reasonable travel lengths. We use the past $1$ second of
observation data as input to predict the bounding boxes of 
vehicles for the next $1$ second. We randomly split the training
($70\%$) and testing  ($30\%$) videos,
resulting in $\sim40,000$ training and $\sim17,000$ testing samples.

\begin{table}[t]
 \vspace{5pt}
 \setlength{\extrarowheight}{4pt}
 \setlength{\tabcolsep}{3pt}
 \centering
 \caption{Quantitative results of proposed methods and baselines on HEV-I dataset with metrics FDE/ADE/FIOU.}
 \scriptsize{
 \begin{tabular}{l|ccc}
     \toprule
     \textbf{Models}
     & \textbf{Easy Cases} & \textbf{Challenging Cases} & \textbf{All Cases}\\
     \midrule
     Linear & 31.49 / 17.04 / 0.68 & 107.93 / 56.29 / 0.33 & 72.37 / 38.04 / 0.50 \\
     ConstAccel & 20.82 / 13.86 / 0.74 & 90.33 / 49.06 / 0.35 & 58.00 / 28.05 / 0.53  \\
     Conv1D~\cite{Yagi_2018_CVPR} &  18.84 / 12.09 / 0.75 & 37.95 / 20.97 / 0.64 & 29.06 / 16.84 / 0.69 \\
     \midrule
     RNN-ED-X & 23.57 / 11.96 / 0.74 & 43.15 / 22.24 / 0.60 & 34.04 / 17.46 / 0.67 \\
     RNN-ED-XE & 22.28 / 11.60 / 0.74 & 42.27 / 22.39 / 0.61 & 32.97 / 17.37 / 0.67 \\
     RNN-ED-XO & 17.45 / 8.68 / 0.78 & 32.61 / 16.72 / 0.66 & 25.56 / 12.98 / 0.72 \\
     RNN-ED-XOE & \textbf{16.72} / \textbf{8.52} / \textbf{0.80} & \textbf{32.05} / \textbf{16.63} / \textbf{0.66} & \textbf{24.92} / \textbf{12.86} / \textbf{0.73} \\
     \bottomrule
 \end{tabular}}
 \label{tab:results1}
 \vspace{-5pt}
 \end{table}

\begin{table}[t]
 \setlength{\extrarowheight}{4pt}
 \setlength{\tabcolsep}{3pt}
 \centering
 \caption{Quantitative results on KITTI dataset. We compare our best model with baselines for simplicity.}
 \footnotesize{
 \begin{tabular}{l|ccc}
     \toprule
     \textbf{Models}
     & FDE & ADE & FIOU  \\
     \midrule
     Linear & 78.19 & 38.21 & 0.33  \\
     ConstAccel & 55.66 & 25.78 & 0.39 \\
     Conv1D~\cite{Yagi_2018_CVPR} & 44.13 & 24.38 & 0.49  \\
     \midrule
     Ours & \textbf{37.11}  & \textbf{17.88} & \textbf{0.53}\\
     \bottomrule
 \end{tabular}}
 \label{tab:results2}
 \vspace{-5pt}
 \end{table}

 \begin{figure*}[t]
    \vspace{5pt}
    \centering
    \includegraphics[width=0.9\textwidth]{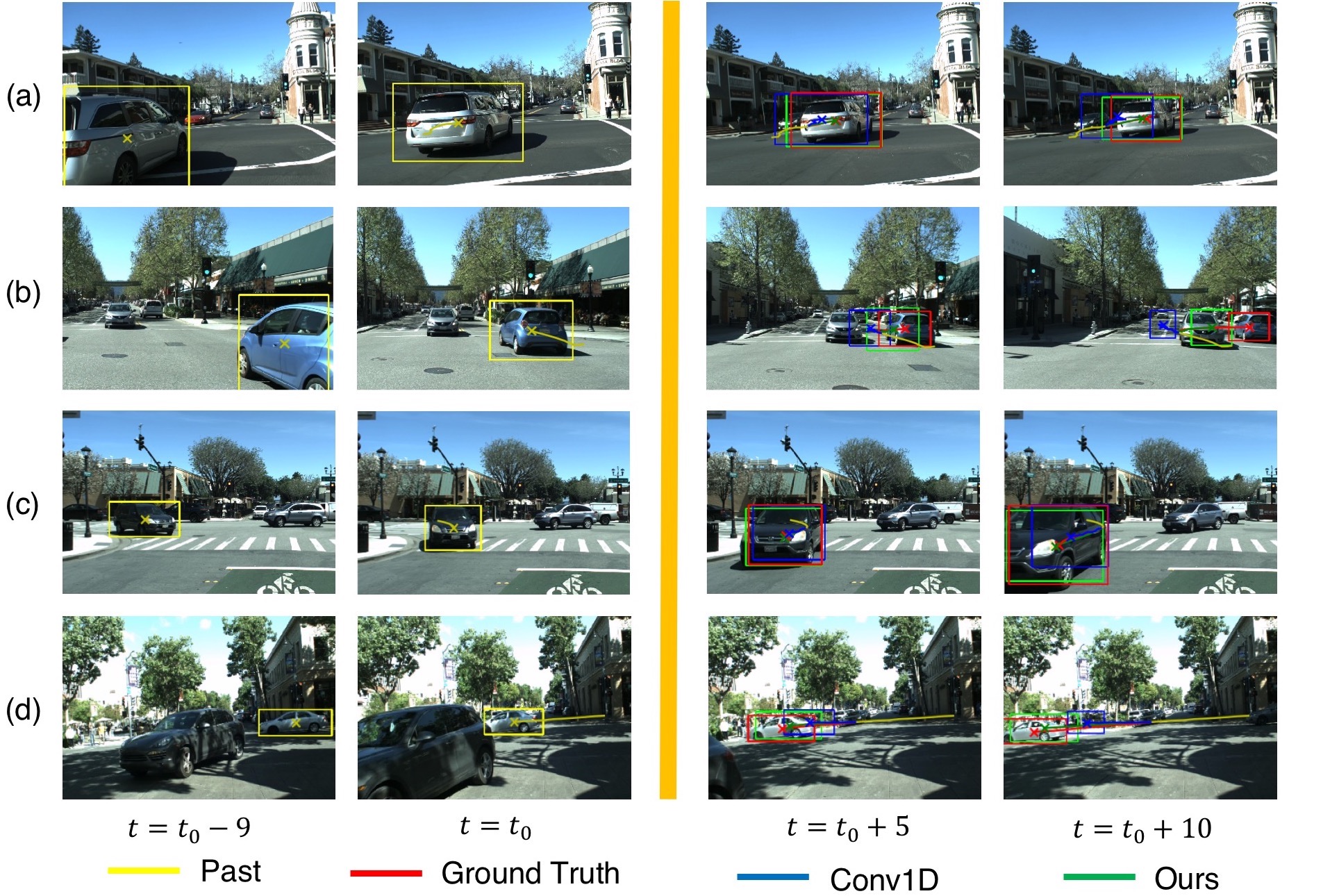}
    \caption{Qualitative results on HEV-I dataset (better in color).}
    \label{fig:qualitative_results}
\end{figure*}

\begin{figure*}[t]
    \centering
    \includegraphics[width=0.9\textwidth]{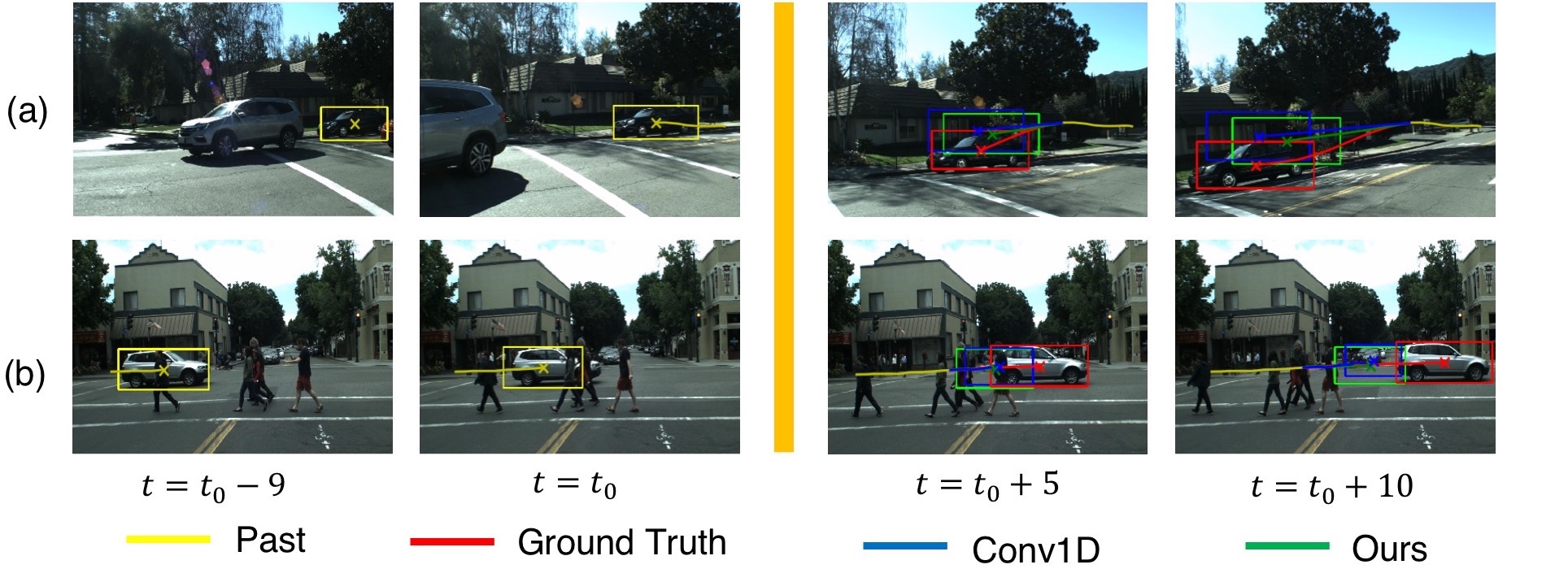}
    \caption{Failure cases on HEV-I dataset (better in color).}
    \label{fig:failure_cases}
    \vspace{-15pt}
\end{figure*}

Statistics of HEV-I are shown in
Fig.~\ref{fig:dataset_dist}. 
As shown, most
vehicle tracklets are short in Fig.~\ref{fig:dataset_dist} (a) because
vehicles usually drive fast and thus leave the field of the first-person view quickly.
Fig.~\ref{fig:dataset_dist} (b) shows the distribution of ego
vehicle yaw angle (in $rad$) across all videos, where positive
indicates turning left and negative indicates 
turning right. It can be seen that HEV-I contains a variety
of different ego motions. Distributions of training and test sample
trajectory lengths (in pixels) are presented in
Fig.~\ref{fig:dataset_dist} (c) and (d). Although most lengths are
shorter than $100$ pixels, the dataset also contains plenty of longer
trajectories. This is important since usually the longer the
trajectory is, the more difficult it is to predict. Compared to
existing data like KITTI, the HEV-I dataset contains more videos and
vehicles, as shown in Table~\ref{tab:datasets}. Most object vehicles in KITTI are parked on the road or driving in the same direction on highways, while in HEV-I, all vehicles
are at intersections and performing diverse maneuvers.

\subsection{Implementation Details}

We compute dense optical flow using
Flownet2.0~\cite{Ilg2017} and use a $5\times 5$ ROIPooling operator
to produce the final flattened feature vector $O_t\in\mathbb{R}^{50}$.
ORB-SLAM2~\cite{MurArtal2017} is used to estimate ego-vehicle motion from
first-person videos.

We use Keras with TensorFlow backend~\cite{abadi2016tensorflow} to implement our model and perform
training and experiments on a system with Nvidia Tesla P100 GPUs. We
use the gated recurrent unit (GRU)~\cite{chung2015gated} as basic RNN cell. Compared to long short-term memory (LSTM)~\cite{hochreiter1997long}, GRU has fewer parameters, which makes it faster without affecting performance~\cite{chung2014empirical}. The hidden state size of our
encoder and decoder GRUs is $512$. We use the Adam~\cite{kingma2014adam} optimizer with fixed learning rate $0.0005$ and batch size $64$. Training is terminated after $40$ epochs and the best models are selected.

\subsection{Baselines and Metrics}
\xhdr{Baselines.} We compare the performance of the proposed method
with several baselines:

\textit{Linear Regression (Linear)} extrapolates future bounding boxes by assuming the location and scale change are linear. 

\textit{Constant Acceleration (ConstAccel)} assumes the object has
constant horizontal and vertical acceleration in the camera frame,
i.e. that the second-order derivatives of $X$ are constant
values.

\textit{Conv1D} is adapted from~\cite{Yagi_2018_CVPR}, by replacing the location-scale and pose
input streams with past bounding boxes and dense optical flow.
%\end{itemize}

To evaluate the contribution of each component of our
model, we also implemented multiple simpler baselines for ablation
studies:

\textit{RNN-ED-X} is an RNN encoder-decoder with only past bounding boxes as inputs.

\textit{RNN-ED-XE} builds on \textit{RNN-ED-X} but also incorporates 
future ego-motion as decoder inputs.

\textit{RNN-ED-XO} is a two-stream RNN encoder-decoder model with past bounding boxes and optical flow as inputs.

\textit{RNN-ED-XOE} is our best model as shown in Fig.\ref{fig:RNN-ED} with awareness of future ego-motion.

\xhdr{Evaluation Metrics.} To evaluate location prediction, we use
final displacement error (FDE)~\cite{Yagi_2018_CVPR} and average
displacement error (ADE)~\cite{Alahi_2016_CVPR}, where ADE emphasizes more on the overall prediction accuracy along the horizon. To evaluate bounding box prediction, we propose a final intersection over union (FIOU) metric that measures overlap between the predicted bounding box and ground truth at the final frame.

\subsection{Results on HEV-I Dataset}
\xhdr{Quantitative Results.}
As shown in Table~\RNum{2}, we split the testing dataset into easy
and challenging cases based on the FDE performance of the
\textit{ConstAccel} baseline. A sample is classified as easy if the
\textit{ConstAccel} achieves FDE lower than the average FDE ($58.00$),
otherwise it is classified as challenging. Intuitively, easy cases
include target vehicles that are stationary or whose future locations
can be easily propagated from the past, while challenging cases
usually involve diverse and intense motion, e.g. the target vehicle
suddenly accelerates or brakes. In evaluation, we report the results of
easy and challenging cases, as well as the overall results on all testing samples.

Our best method (\textit{RNN-ED-XOE}) significantly outperforms naive
baselines including \textit{Linear} and
\textit{ConstAccel} on all cases (FDE of
\textbf{24.92} vs. 72.37 vs. 58.00). It also improves about $15\%$ from
the state-of-the-art \textit{Conv1D} baseline. The improvement on challenging
cases is more significant since future trajectories are complex and
temporal modeling is more difficult. To more fairly compare the
capability of RNN-ED and convolution-deconvolution models, we compare \textit{RNN-ED-XO}
with \textit{Conv1D}. These two methods use the same features as
inputs to predict future vehicle bounding boxes, but rely on
different temporal modeling frameworks. The results (FDE of
\textbf{25.56} vs 29.06) suggest that the RNN-ED architecture offers better
temporal modeling compared to \textit{Conv1D}, because the
convolution-deconvolution model generates future trajectory in one
shot while the RNN-ED model generates a new prediction based
on the previous hidden state. Ablation
studies also show that dense optical flow features are essential to
accurate prediction of future bounding boxes, especially for
challenging cases. The FDE is reduced from $34.04$ to $25.56$ by
adding optical flow stream (\textit{RNN-ED-XO}) to \textit{RNN-ED-X} model. By using future
ego-motion, performance can be further improved as shown in the last
row of Table \RNum{2}.

%\ref{tab:results}.

\xhdr{Qualitative Results.}
Fig.~\ref{fig:qualitative_results} shows four sample results of our
best model (in green) and the \textit{Conv1D} baseline (in blue). Each row
represents one test sample and each column corresponds to each time
step. The past and prediction views are separated by the yellow
vertical line. Example (a) shows a case where the initial bounding
box is noisy because it is close to the image boundary, and our
results are more accurate than those of \textit{Conv1D}.
Example (b) shows how our model, with awareness of future ego-motion,
can predict object future location more
accurately while the baseline model predicts future location in the
wrong direction. Examples (c) and (d) show that for a curved or long
trajectory, our model provides better temporal modelling than \textit{Conv1D}.
These results are consistent with our evaluation observations.

\xhdr{Failure Cases.} Although our proposed method generally performs
well, there are still limitations. Fig.\ref{fig:failure_cases} (a)
shows a case when the ground truth future path is curved due to uneven
road surface, which our method fails to consider.  In
Fig.\ref{fig:failure_cases} (b), the target vehicle is occluded by
pedestrians moving in the opposite direction, which creates misleading
optical flow that leads to an inaccurate bounding box (especially in
$t=t_0$ frame). 
Future work could avoid this type of error by better modeling the
entire traffic scene as well as relations between traffic participants.

\subsection{Results on KITTI Dataset}
We also evaluate our method on a 38-video subset of the KITTI raw dataset,
including city, road and residential
scenarios.
Compared to HEV-I, the road surface of KITTI is more uneven and vehicles are mostly parked on the side of the road with occlusions. Another difference is that in HEV-I, the ego-vehicle often stops at intersections to yield to other vehicles, resulting in static samples with no motion at all. We did not remove static samples from the dataset since predicting a static object is also valuable.

To evaluate our method on KITTI, we first generate the input features following the same process of HEV-I dataset, resulting in
$\sim 8000$ training and $\sim 2700$ testing samples. Performance of baselines
and our best model are shown in Table \RNum{3}. Both learning-based models are trained for 40 epoches and the best models are selected.  The results show that our
method outperforms all baselines including the state-of-the-art
Conv1D (FDE of \textbf{37.11} vs 78.19 vs 55.66 vs 44.13). We also observe that both learning-based methods did not perform as well as they did on HEV-I. One possible reason is that KITTI is much smaller so that the models are not fully trained. In general, we conclude that the use of the proposed framework results in more robust future vehicle localization across different datasets.

%% file: Conclusion.tex
\section{CONCLUSION}\label{sec:conclusion}
We proposed the new problem of predicting the relative
location and scale of target vehicles in first-person video. We presented a new
dataset  collected from intersection scenarios to include as
many vehicles and motion as possible. Our proposed multi-stream RNN
encoder-decoder structure with awareness of future ego motion shows
promising results compared to other baselines on our dataset as well as 
on KITTI, and we tested how each component contributed to the model
through an ablation study.

Future work includes incorporating evidence from scene context,
traffic signs/signals, depth data, and other
vehicle-environment interactions. Social relationships such as
vehicle-to-vehicle and vehicle-to-pedestrian interactions could also
be considered. 